\documentclass{article}
\usepackage{spconf,amsmath,graphicx}
\usepackage{amssymb}
\usepackage{booktabs}
\usepackage{subcaption}
\usepackage{graphicx}
\usepackage{multirow}
\usepackage{xcolor}
\usepackage{color, colortbl}

\usepackage{xspace}
\makeatletter
\DeclareRobustCommand\onedot{\futurelet\@let@token\@onedot}
\def\@onedot{\ifx\@let@token.\else.\null\fi\xspace}

\makeatother

\definecolor{Gray}{gray}{0.9}
\definecolor{color1}{HTML}{ECF4F9}
\definecolor{color2}{HTML}{FFF1E0}
\definecolor{color3}{HTML}{ECF4E9}

\title{Cross Pseudo-Labeling for Semi-Supervised \\ Audio-Visual Source Localization}
\name{Yuxin Guo, Shijie Ma, Yuhao Zhao, Hu Su, Wei Zou*\thanks{* corresponding author}}
\address{School of Artificial Intelligence, University of Chinese Academy of Sciences\\
State Key Laboratory of Multimodal Artificial Intelligence Systems (MAIS),\\
Institute of Automation of Chinese Academy of Sciences}
\begin{document}
%\ninept
%
\maketitle
\begin{abstract}

Audio-Visual Source Localization (AVSL) is the task of identifying specific sounding objects in the scene given audio cues. In our work, we focus on semi-supervised AVSL with pseudo-labeling. To address the issues with vanilla hard pseudo-labels including bias accumulation, noise sensitivity, and instability, we propose a novel method named \textbf{Cross Pseudo-Labeling (XPL)}, wherein two models learn from each other with the cross-refine mechanism to avoid bias accumulation. We equip XPL with two effective components. Firstly, the soft pseudo-labels with sharpening and pseudo-label exponential moving average mechanisms enable models to achieve gradual self-improvement and ensure stable training. Secondly, the curriculum data selection module adaptively selects pseudo-labels with high quality during training to mitigate potential bias. Experimental results demonstrate that XPL significantly outperforms existing methods, achieving state-of-the-art performance while effectively mitigating confirmation bias and ensuring training stability.

\end{abstract}
\begin{keywords}
Audio-Visual Source Localization, Audio-Visual Learning, Semi-Supervised Learning, Pseudo-Labeling. 
\end{keywords}
\section{Introduction}
\label{sec:intro}

When we hear a sound, the combination of vision and audition enables us to locate the specific source of the sound within a scene. Audio-Visual Source Localization (AVSL) aligns visual and auditory information which benefits downstream tasks like navigation~\cite{chen2020soundspaces}, sound source separation~\cite{majumder2022active}, and enhances fine-grained multimodal pre-training.

Existing methods predominantly rely on self-supervised contrastive learning for audio-visual correspondence. Attention10k~\cite{senocak2018learning, senocak2019learning} employs an attention-based dual-stream network for localization, DMC~\cite{hu2019deep} establishes audio-visual clustering, LVS~\cite{chen2021localizing} introduces a tri-map for foreground-background separation, EZVSL~\cite{mo2022localizing} utilizes global max pooling for focused alignment, LCBM~\cite{senocak2022learning} employs a simple audio-visual classification model, and HardPos~\cite{senocak2022learning} mines semantically similar samples for contrastive learning. Additionally, various studies have tackled specific challenges, like false positives~\cite{senocak2022learning,mo2022SLAVC, liu2022visual}, false negatives~\cite{park2023marginnce, sun2023learning}, and multi-source issues~\cite{hu2020discriminative, Hu_Wei_Qian_Lin_Song_Wen, qian2020multiple, hu2022mix, mo2023audio}. However, in principle, self-supervised AVSL is not a fully learnable task due to the absence of positional labels, leading to suboptimal performance, including inaccurate localization, excessive foreground, inability to pinpoint specific locations, and a failure to achieve precise object-level localization~\cite{guo2023dual}.

Attention10k-SSL~\cite{senocak2018learning, senocak2019learning} is a pioneer work to introduce semi-supervised learning (SSL) to AVSL with positional labels, however, it suffers from overfitting and poor generalization due to exclusive reliance on labeled data without harnessing the rich information in unlabeled data. On the other hand, pseudo-labeling~\cite{lee2013pseudo}, as an advanced SSL technique, is highly effective in feature extraction and well-suited for AVSL. Despite its strong performance, vanilla hard pseudo-labeling (PL) faces three main challenges when directly applied to localization tasks: 
1). The initial pseudo-labels are prone to inaccuracy and cannot be rectified by a single model, resulting in confirmation bias~\cite{arazo2020pseudo}.
2). One-hot hard labels exhibit heightened sensitivity to noise owing to their overconfident nature.
3). Vanilla pseudo-labels tend to oscillate and bring about instability in training.

\begin{figure*}[!t]
    \centering
    \includegraphics[width=.98\textwidth]{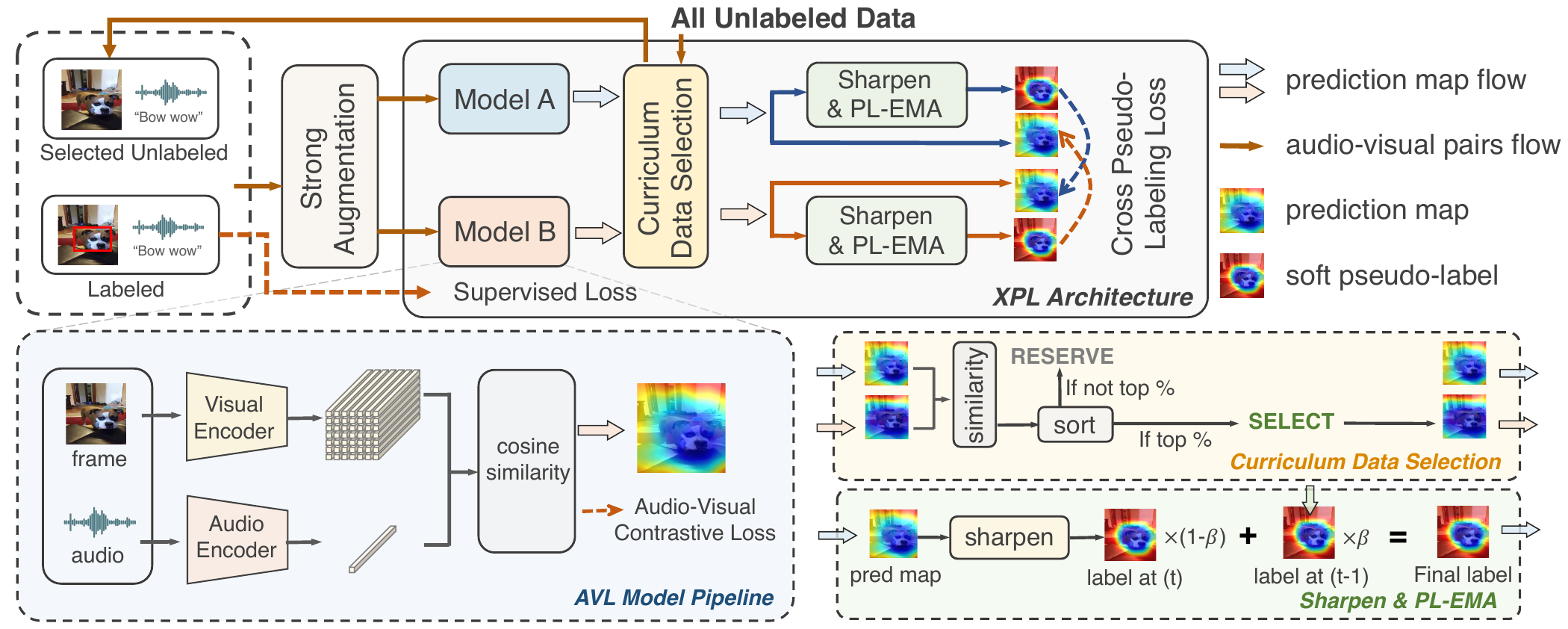}
    \vspace{-10pt}
    \caption{An overview of the proposed XPL (\textbf{up}). Two distinct AVSL models (\textbf{bottom blue}) generate prediction maps for input audio-visual pairs. To ensure stability by an initial, Curriculum Data Selection mechanism (\textbf{bottom yellow}) sorts the samples by reliability and feeds them to the model in batches. Then, Sharpening and PL-EMA module (\textbf{bottom green}) sharpens the prediction map, and performs exponential moving average (EMA) with the previous time step's pseudo-label to obtain the final pseudo-label. Finally, each model utilizes the pseudo-labels generated by the other for training and thus rectifying bias.}
    \label{fig:xpl-framework}
    \vspace{-10pt}
\end{figure*}

To address these issues, we propose a cross pseudo-labeling semi-supervised method, called XPL, with a cross-refine mechanism and a novel soft pseudo-labeling mechanism. Specifically, XPL trains two separate backbone models individually, with each model using pseudo-labels generated by the other as supervision to mitigate bias from different perspectives. We use soft pseudo-labels and a pixel-wise sharpening mechanism to address the overconfidence of one-hot hard labels, promoting self-training by gradually generating more confident outputs while riching information. Additionally, a pseudo-label exponential moving average (PL-EMA) module is proposed to mitigate instability caused by oscillating pseudo-labels, ensuring stability during self-training. Additionally, to ensure initial pseudo-label accuracy, we design a Curriculum Data Selection mechanism to gradually select reliable samples and alleviate confirmation bias issue.

To sum up, our contributions encompass three aspects:

\begin{itemize}
    \vspace{-5pt}
    \item We propose a novel semi-supervised AVSL method with a cross-refine and a curriculum data selection mechanism, which trains models from different perspectives and effectively mitigates confirmation bias in pseudo-labeling of semi-supervised learning.
    \vspace{-7pt}
    \item We design a pixel-wise soft pseudo-labeling mechanism, combining sharpening and EMA techniques to retain information from pseudo-labels, significantly enhancing training stability and self-improvement.
    \vspace{-7pt}
    \item Our method significantly improves localization accuracy across various datasets, achieving state-of-the-art performance and effectively addressing issues like poor generalization and unstable training.
\end{itemize}

\vspace{-7pt}
\section{Methodology} \label{sec:methods}

We provide an overview of the architecture of our method in Fig.~\ref{fig:xpl-framework}, comprising Curriculum Data Selection (Section~\ref{sec:data}), Sharpening and PL-EMA module (Section~\ref{sec:plema}), and the Cross-Refine mechanism (Section~\ref{sec:cross-refine}). We also summarize our training objectives in Section~\ref{sec:loss}.

\textbf{Notations.}
The image frame $v_i$ and the corresponding audio $a_i$ from the $i$-th audio-visual pair $\{v_i, a_i\}$, and the visual feature $f(v_i)$ and the audio feature $g(a_i)$ are fed into the localization models to get the prediction map $\mathcal{M}_{i}\in \mathbb{R}^{1\times H\times W} $. $\mathcal{G}_{i}\in \mathbb{R}^{1\times H\times W} $ and $\mathcal{PL}_{i}\in \mathbb{R}^{1\times H\times W} $ are ground-truth and pseudo labels. $\mathcal{D}_l$ and $\mathcal{D}_u$ denote labeled and unlabeled datasets. A and B represent two separate AVSL models with different backbones, respectively.

% \subsection{Overview of Architecture}
% \label{sec:overview}
% As in Fig.~\ref{fig:xpl-framework}, two distinct AVSL models with different backbones generate prediction maps for input audio-visual pairs. To ensure stability by an initial, we sort the samples by reliability and feed them to the model in batches. Then, we sharpen the map of selected data, and perform exponential moving average (EMA) with the previous time step's pseudo-labels to obtain the final soft pseudo-labels.  Finally, each model utilizes the pseudo-labels generated by the other for training and rectifying potential bias.

\subsection{Curriculum Data Selection}
\label{sec:data}

Both model. A and model. B initially are trained on the labeled data for better pseudo-labels. To maintain training stability, we employ curriculum learning by gradually introducing more samples ranked by reliability. Specifically, a sample is considered reliable when its prediction maps from two models reach a consensus. We thus employ the Pearson correlation coefficient $\rho_{i}$ of the two prediction maps $\mathcal{M}_i^{k} (k=A,B)$ of the $i$-th sample to measure the consensus:
\vspace{-5pt}
\begin{align}
\small
    \mathcal{M}_i^k= \mathrm{sim}(g^k(a_i),f^k(v_i))=\frac{\left \langle g^k(a_i) ,  f^k(v_i) \right \rangle }{\left \| g^k(a_i) \right \| \cdot \left \| f^k(v_i)\right \| }, \label{eq_sim} \\
    \rho_{i}=\frac{\mathbb{E}\left[\left(\mathcal{M}^{A}_{i}-\mu_{\mathcal{M}^{A}_{i}}\right)\left(\mathcal{M}^{B}_{i}-\mu_{\mathcal{M}^{B}_{i}}\right)\right]}{\sigma_{\mathcal{M}^{A}_{i}} \cdot \sigma_{\mathcal{M}^{B}_{i}}}, \label{pearson} \\
    \mathcal{D}_{all} = \mathcal{D}_{l} \cup \Big\{(a_i,v_i)\ \big\vert\ \rho_i > \delta,\ \forall(a_i,v_i)\in \mathcal{D}_{u}\Big\}.
\end{align}
Where the mean $\mu$ and variance $\sigma$ are computed across the spatial-dimension of maps. We begin training with the most reliable samples and gradually involve more data in a ramp-up way. Samples with $\rho \le 0.8$ are rejected due to high instability, ensuring both model accuracy and stability.

\vspace{-7pt}
\subsection{Soft Pseudo-Label with Sharpening and PL-EMA}
\label{sec:plema}
% refine
After obtaining the prediction map of each model, we pass it through a sigmoid function $\varphi(x)$ (shifted by $0.5$), which is an S-shaped curve that can make the prediction scores of pixels greater than 0.5 closer to 1 and those less than 0.5 closer to 0, to sharpen and serves as the instantaneous soft pseudo-label $\mathcal{\widehat{PL}}$ generated by the model:
\begin{align}
    \mathcal{\widehat{PL}}^t_i =  {\varphi }  (\mathcal{M}^t_i), \ \text{where} \ {\varphi}(x)= 1 + \frac{1}{1+e^{-a(x-0.5)}}. 
\end{align}
Where $a$ controls the smoothness of the sharpening. Afterward, we store the historical pseudo-labels in a memory bank and apply the PL-EMA to yield the final soft pseudo-label at time $t$:
\begin{align}
    \mathcal{PL}^{k,t}_i =\beta \mathcal{PL}^{t-1}_i +(1-\beta)\mathcal{\widehat{PL}}^t_i.
\end{align}
Where $\beta$ denotes the updating rate of PL-EMA. Sharpening boosts model confidence for self-improvement, while PL-EMA ensures training stability by preventing pseudo-label oscillations from model updates and noise.

\subsection{Cross-Refine Mechanism}
\label{sec:cross-refine}
In pseudo-label methods, a single model cannot rectify bias during training. Therefore, we train two models individually, using each other's pseudo-labels as their own supervision, with a loss $\mathcal{L_\text{cross}}$ based on pixel-wise cross-entropy $H(\cdot,\cdot)$:
\begin{align}
    \mathcal{L}_\text{cross}= \mathbb{E}_{i\sim \mathcal{D}_{all}} {H}(\mathcal{PL}_{i}, \mathcal{M}_i). \label{eq_suploss}
\end{align}
Where $\mathcal{PL}$ and $\mathcal{M}$ are from two different models. To ensure that the models are trained from different perspectives, we employ distinct backbones and augmentations for the two models. This mechanism enables the model to continually generate higher-quality pseudo-labels and effectively mitigate confirmation bias.

\subsection{Training Objectives}
\label{sec:loss}
\noindent\textbf{Cross Pseudo-labeling Loss.} We utilize the generated soft pseudo-labels as supervision signals to train another model, as detailed in Section~\ref{sec:cross-refine}.

\noindent\textbf{Supervised Loss.} To ensure training stability and prevent model collapse, we maintain the supervised loss of the labeled data throughout the training process:
\begin{align}
    \mathcal{L}_\text{sup}= \mathbb{E}_{i\sim \mathcal{D}_l} H(\mathcal{G}_{i}, \mathcal{M}_i). \label{eq_suploss}
\end{align}

\noindent\textbf{Audio-Visual Contrastive Loss.} To mitigate the semantic gap between visual and auditory features and align them, we introduce the audio-visual contrastive loss, which employs global max pooling of visual features and the info-NCE loss:
\begin{equation}
    \small
    \begin{aligned}
        \mathcal{L}_\text{unsup}=-\mathbb{E}_{(a_i, v_i)\sim \mathcal{D}_{all}}\Big[\log \frac{\exp (s(A_i, V_i) / \tau)}{\sum_{j=1}^{n} \exp \left(s\left(A_i, V_j)\right) / \tau\right)} &\\
        +\log \frac{\exp (s(V_i, A_i) / \tau)}{\sum_{j=1}^{n} \exp \left(s\left(V_i, A_j\right) / \tau\right)}\Big]. &\label{eq:loss-unsup}
    \end{aligned}
\end{equation}
Where $A_i=g(a_i), V_j=GMP(f(v_j))$, `s' represents cosine-similarity, and `$\tau$' is the temperature parameter. Overall, the learning objective for training is:
% \vspace{-2pt}
\begin{align}
    \mathcal{L}_\text{full}=\sum_{A, B} (\mathcal{L}_\text{cross}+\mathcal{L}_\text{sup}+\lambda_{u}\mathcal{L}_\text{unsup}).
\end{align}
Where $\lambda_u$ is set to be 0.5, and A, B represent two backbones.

\begin{table}[!t]
\setlength\tabcolsep{4pt}
\centering
\renewcommand{\arraystretch}{1}
\caption{Comparison results with \colorbox{color1}{self-supervised}, \colorbox{color2}{supervised} and \colorbox{color3}{semi-supervised} AVSL methods.}
\vspace{-9pt}
\label{tab:main}
\resizebox{1\linewidth}{!}{
\begin{tabular}{@{}ccccccc@{}}
\toprule
\multirow{2}{*}{Methods} & \multicolumn{2}{c}{Flickr-144k} & \multicolumn{2}{c}{Vggss-144k} & \multicolumn{2}{c}{Cross-dataset} \\ \cmidrule(l){2-3} \cmidrule(l){4-5} \cmidrule(l){6-7}
 & CIoU $\uparrow$ & AUC $\uparrow$ & CIoU $\uparrow$ & AUC $\uparrow$ & CIoU $\uparrow$ & AUC $\uparrow$ \\ \midrule
\cellcolor{color1}Attention10k~\cite{senocak2018learning,senocak2019learning} & 66.00 & 55.80 & 18.50 & 30.20 & 66.00 & 55.80 \\
\cellcolor{color1}LCBM~\cite{senocak2022less} & 68.82 & 56.68 & 32.20 & 36.60 & 69.64 & 57.02 \\
\cellcolor{color1}LVS~\cite{chen2021localizing} & 69.60 & 57.30 & 34.40 & 38.20 & 71.90 & 58.20 \\
\cellcolor{color1}HardPos~\cite{senocak2022learning} & 76.80 & 59.20 & 34.60 & 38.00 & 76.80 & 59.20 \\
\cellcolor{color1}EZVSL~\cite{mo2022localizing} & 72.69 & 58.70 & 34.38 & 37.70 & 79.51 & 61.17 \\
\cellcolor{color1}SLAVC~\cite{mo2022SLAVC} & 73.84 & 58.98 & 39.20 & 39.46 & 80.00 & 61.68 \\
\cellcolor{color1}SSL-TIE~\cite{liu2022exploiting} & 81.50 & 61.10 & 38.60 & 39.60 & 79.50 & 61.20 \\
\cellcolor{color1}MarginNCE~\cite{park2023marginnce} & 83.94 & 63.20 & 38.25 & 39.06 & 83.94 & 63.20 \\
\cellcolor{color1}FNAC~\cite{sun2023learning} & 78.71 & 59.33 & 39.50 & 39.66 & 84.73 & 63.76 \\ \hline 
\cellcolor{color2}Sup only & 84.68 & 62.76 & 42.68 & 41.06 & 80.82 & 61.74 \\ \hline
\cellcolor{color3}Attention10k-SSL~\cite{senocak2018learning,senocak2019learning} & 84.40 & 62.41 & 43.26 & 41.68 & 81.16 & 61.92 \\
\cellcolor{color3}Vanilla hard PL & 82.36 & 62.20 & 40.68 & 40.12 & 80.14 & 61.86 \\ \hline \hline
\cellcolor{color3}XPL (ours) & \textbf{89.28} & \textbf{68.24} & \textbf{48.40} & \textbf{44.12} & \textbf{88.04} & \textbf{66.86} \\ \bottomrule
\end{tabular}
}
\vspace{-10pt}
\end{table}

\section{Experiment}
\label{sec:exp}

\begin{figure*}[!t]
    \centering
    \includegraphics[width=.98\textwidth]{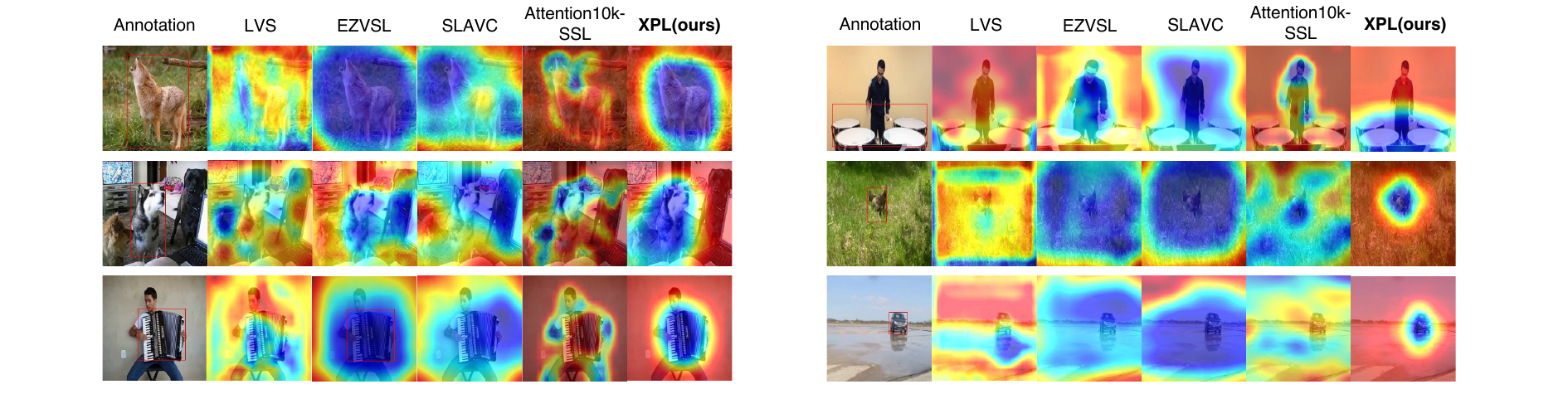}
    \vspace{-10pt}
    \caption{Visualization of the proposed XPL. We can observe that XPL can localize sounding objects of various sizes accurately and effectively distinguishing foreground and background elements.}
    \label{fig:viz}
\end{figure*}

\newcommand{\components}{
\begin{tabular}{@{}ccc@{}}
\toprule
 & CIoU $\uparrow$ & AUC $\uparrow$ \\ \midrule
w/ vanilla hard PL & 82.36 & 62.20 \\
w/o PL-EMA & 84.26 & 64.54 \\
w/o Cross-Refine & 85.38 & 65.42 \\
w/o Sharpening & 86.74 & 66.26 \\
w/o Data Selection & 87.60 & 66.91 \\ \midrule
XPL (ours) & \textbf{89.28} & \textbf{68.24} \\ \bottomrule
\end{tabular}
}

\newcommand{\ema}{
\begin{tabular}{@{}ccc@{}}
\toprule
$\beta$ & CIoU $\uparrow$ & AUC $\uparrow$ \\ \midrule
0.1 & 85.22 & 65.54 \\
0.3 & 85.38 & 66.32 \\
0.5 & 87.80 & 67.58 \\
\textbf{0.7} & \textbf{89.28} &\textbf{ 68.24} \\
0.9 & 88.64 & 67.92 \\ \bottomrule
\end{tabular}
}

\newcommand{\openset}{
\begin{tabular}{@{}ccc@{}}
\toprule
                 & CIoU $\uparrow$  & AUC $\uparrow$   \\ \midrule
EZVSL~\cite{mo2022localizing}            & 32.66 & 36.72 \\
MarginNCE~\cite{park2023marginnce}        & 37.90 & 39.17 \\ 
Attention10k-SSL~\cite{senocak2018learning,senocak2019learning} & 20.32 & 26.58 \\
vanilla hard PL       & 19.22 & 24.26 \\ \midrule
XPL(ours)              & \textbf{43.08} & \textbf{42.84} \\ \bottomrule
\end{tabular}
}

\newcommand{\stable}{
    \includegraphics[width=.98\textwidth]{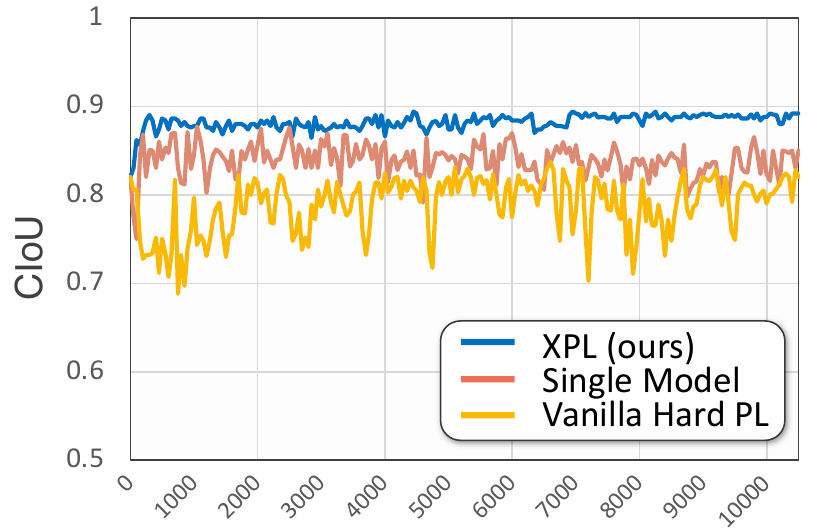}
    \vspace{-10pt}
    \label{fig:stable}
}

% \begin{table*}[!t]
%     \setlength\tabcolsep{6pt}
%     \centering
%     \renewcommand{\arraystretch}{0.8}
%     \vspace{-5pt}
%     \begin{subfigure}{0.305\linewidth}
%         \resizebox{\linewidth}{!}{\openset}
%         \caption{Openset}
%         \label{tab:ablation-openset}
%     \end{subfigure}
%     \begin{subfigure}{0.26\linewidth}
%         \resizebox{\linewidth}{!}{\components}
%         \caption{Effect of each component.}
%         \label{tab:ablation-components}
%     \end{subfigure}
%     \begin{subfigure}{0.195\linewidth}
%         \resizebox{\linewidth}{!}{\ema}
%         \caption{$\beta$ in EMA.}
%         \label{tab:ablation-ema}
%     \end{subfigure}
%     \vspace{-8pt}
%     \caption{Ablation studies.}
%     \label{tab:ablations}
%     \vspace{-10pt}
% \end{table*}

\begin{table*}[!t]
    \setlength\tabcolsep{11pt}
    \centering
    \renewcommand{\arraystretch}{1}
    \vspace{-5pt}
    \resizebox{.98\textwidth}{!}{
    \begin{subfigure}{0.43\linewidth}
        \resizebox{.91\linewidth}{!}{\openset}
        \caption{Openset}
        \label{tab:ablation-openset}
    \end{subfigure}\hfill
    \begin{subfigure}{0.345\linewidth}
        \resizebox{.9\linewidth}{!}{\components}
        \caption{Effect of each component.}
        \label{tab:ablation-components}
    \end{subfigure}\hfill
    \begin{subfigure}{0.28\linewidth}
        \resizebox{.89\linewidth}{!}{\ema}
        \caption{$\beta$ in EMA.}
        \label{tab:ablation-ema}
    \end{subfigure} \hfill
    \begin{subfigure}{0.33\linewidth}
        \resizebox{.89\linewidth}{!}{\stable}
        \caption{Model performance.}
        \label{tab:ablation-ciou}
    \end{subfigure}
    }
    \vspace{-8pt}
    \caption{Ablation study results. Exp. (a) is trained on Vggss-144k while Exp. (b)-(d) are trained on Flickr-144k.}
    \label{tab:ablations}
    \vspace{-10pt}
\end{table*}

\subsection{Experimental Settings}
\vspace{-5pt}
\textbf{For datasets,} we conduct experiments on Flickr-SoundNet~\cite{senocak2018learning} and VGG-SoundSource~\cite{Chen20} datasets, using subsets of 10k and 144k as unlabeled training sets with 2.5k labeled data. \textbf{For backbones}, we employ ResNet-18~\cite{he2016deep} as the image encoder, while Vggish~\cite{hershey2017cnn} and SoundNet~\cite{aytar2016soundnet} as audio encoders. They all loaded pre-trained parameters. \textbf{For augmentation}, we use color transformations in RandAugment~\cite{cubuk2020randaugment}. To ensure high data-diversity, we apply distinct augmentations for the two pipelines via randomness. \textbf{For metrics,} we use CIoU and AUC. \textbf{For comparative methods,} we select some self-supervised methods~\cite{senocak2018learning,senocak2019learning,chen2021localizing,mo2022localizing,senocak2022learning,liu2022exploiting,mo2022SLAVC,park2023marginnce,sun2023learning,senocak2022less} and semi-supervised method Attention10k-SSL~\cite{senocak2018learning,senocak2019learning}. We also compare the performance of our method against supervised-only and vanilla hard pseudo-label baselines.

\vspace{-7pt}
\subsection{Comparison with SOTA}
\vspace{-5pt}
\noindent\textbf{Localization Performance.}
We compare XPL with existing methods on both datasets. From Table~\ref{tab:main}, one can observe that XPL outperforms self-supervised methods by $5.34\%, 8.90\%$ and semi-supervised Attention10k-SSL by $4.88\%$ and $5.14\%$. Visualizations in Fig.~\ref{fig:viz} demonstrate that XPL excels at accurately locating object-level sources and effectively distinguishing between foreground and background, which can be attributed to the stable sharpening soft pseudo-label and cross-refine mechanism.

\noindent\textbf{Generalization Capability.}
The cross-dataset results in Table~\ref{tab:main} reveal that Attention10k-SSL overfits towards specific datasets, leading to $3.24\%$ performance decrease, while XPL maintains advanced performance. Another assessment we conduct to evaluate the model's generalization capability is in open-set settings~\cite{mo2022SLAVC, ma2023towards, zhu2024openworld}, where testing samples come from categories not used during training. Results in Table~\ref{tab:ablation-openset} indicate that XPL continues to outperform Attention10k-SSL and other methods in open-set environments, confirming the generalization capabilities of XPL. This can be attributed to accurately extracting local features for unlabeled data through soft pseudo-labels, which encompass rich information and thus mitigate the overfitting issue.

\noindent\textbf{Stability.}
XPL significantly outperforms the vanilla hard pseudo-label baseline by $6.92\%, 7.72\%$, as shown in Table~\ref{tab:main}. Additionally, we also demonstrate the training stability of XPL and vanilla hard PL. Fig.2(d) highlights that the vanilla PL (yellow curve) exhibits high instability and variance, while XPL (blue curve) maintains training stability and steadily improves the performance due to the proposed soft pseudo-label with sharpening and PL-EMA mechanisms.

% The results are shown in Table~\ref{tab:ablations}. 

% ablation results
\vspace{-7pt}
\subsection{Ablation Study}
\label{sec:ablation}
\vspace{-2pt}

\noindent\textbf{Module Effectiveness.}
To validate the effectiveness of each module, we conduct ablation experiments on Sharpening, PL-EMA, Curriculum Data Selection, and Cross-Refine mechanism. Table~\ref{tab:ablation-components} highlights that the hard pseudo-label performs worse than the soft one due to overconfidence. Additionally, without any of these modules will lead to instability or bias accumulation and thus performance decreases, which emphasizes the indispensability of each module in XPL.

% \noindent\textbf{Label Sharpening Mechanism.}
% We explored sigmoid functions with various parameters. Specifically, we investigated the impact of different sigmoid functions' steepness and inflection point values on training. As shown in Table n-(a)-(d), increasing the value of 'a' in the sigmoid function results in a steeper slope near the inflection point. We found that a steeper slope makes the sharpened pseudo-labels closer to hard pseudo-labels, leading to information loss and overly confident predictions, which can accumulate bias and impact model training performance. We chose 'a' to be 20.

\noindent\textbf{EMA rate.}
To understand how the EMA rate $\beta$ of PL-EMA affects model training, we conducted experiments with different values of $\beta$. 
% We observed the convergence rate, stability, and final performance of the model. 
As shown in Table~\ref{tab:ablation-ema}, higher $\beta$ results in slower updates, making it difficult to correct inaccurate initial pseudo-labels, while lower $\beta$ leads to faster updates and increasing instability. We finally choose $\beta = 0.7$ as a trade-off.

\noindent\textbf{Cross-Refine and Confirmation Bias.}
We evaluated the ability to mitigate the confirmation bias of Cross-Refine mechanism. As observed in Table~\ref{tab:ablation-components} and Fig. 2(d), the model with cross-refine achieves better performance, while without it, the model accumulates bias over time and cannot correct itself, leading to a continuous decline in performance, which highlights the effective mitigation of confirmation bias.

\vspace{-7pt}
\subsection{Superiority Analysis}
\label{sec:further}
\vspace{-2pt}
% The superiority can be attributed to two major improvements.
\noindent\textbf{Effectively mitigates confirmation bias.}
When the initial pseudo-labels are incorrect, the model will continually adapt to these biased labels, thereby compromising the model's accuracy. The Cross-Refine mechanism helps the two models learn from different perspectives and mutually correct bias in each other. Moreover, the ramp-up scheme ensures model confidence at the initial stages. These two modules enable XPL to effectively address confirmation bias, thereby continually boosting the model's performance.

\noindent\textbf{Significantly ensures training stability.}
Oscillating hard pseudo-labels will significantly impact the stability of the model. The soft pseudo-label focuses on object-level features, enriching the information for learning and ensuring stability. PL-EMA effectively prevents pseudo-label oscillation caused by model variations. Continuous sharpening of pseudo-labels facilitates effective self-training, leading to a consistent improvement in performance.

% \section{Further Discussion}
% \label{sec:discuss}
\vspace{-7pt}
\section{Conclusion}
\label{sec:conclusion}
\vspace{-5pt}

In this paper, we address the limitations in existing self-supervised and semi-supervised AVSL methods. We introduce a novel semi-supervised approach based on pseudo-labels, significantly enhancing localization performance. We tackle challenges like inaccuracy and training instability by soft pseudo-labeling with sharpening and PL-EMA, and address confirmation bias using a cross-refine mechanism. These enhancements improve AVSL while highlighting the importance of annotations and local features. Overall, we hope our approach to inspire research in audio-visual tasks like segmentation, sound source separation, and navigation.

\vfill\pagebreak

\fontsize{9pt}{11pt}\selectfont

% \section{REFERENCES}
% \label{sec:refs}

% List and number all bibliographical references at the end of the
% paper. The references can be numbered in alphabetic order or in
% order of appearance in the document. When referring to them in
% the text, type the corresponding reference number in square
% brackets as shown at the end of this sentence \cite{C2}. An
% additional final page (the fifth page, in most cases) is
% allowed, but must contain only references to the prior
% literature.

\bibliographystyle{IEEEbib}
% \bibliography{strings,refs}
\bibliography{refs}

\end{document}